\documentclass{amia}
\usepackage{graphicx}
\usepackage[labelfont=bf]{caption}
\usepackage[superscript,nomove]{cite}
\usepackage[symbol]{footmisc}
\usepackage{color} 
\usepackage{wrapfig}
\graphicspath{ {./figures/}}
\usepackage{xurl}
\usepackage[hidelinks]{hyperref}

\begin{document}
\title{Launching into clinical space with medspaCy:\\ a new clinical text processing toolkit in Python }
\author {Hannah Eyre, MS$^{{1},{2}}$, Alec B Chapman, MS$^{{1},{2}}$, Kelly S Peterson, MS$^{{2},{3}}$, \\ Jianlin Shi, MD, PhD$^{2}$, Patrick R Alba, MS$^{{1},{2}}$, Makoto M Jones, MD$^{{1},{2}}$,\\ Tamára L Box, PhD$^{{3}}$, 
Scott L DuVall, PhD$^{{1},{2}}$, Olga V Patterson, PhD$^{{1},{2}}$}
\institutes{
  $^1$VA Salt Lake City Health Care System;
  $^2$University of Utah, Salt Lake City, UT, USA;
  $^3$Veterans Health Administration Office of Analytics and Performance Integration
}
\maketitle
\renewcommand*\footnoterule{}
\noindent{\bf Abstract}  
\textit{Despite impressive success of machine learning algorithms in clinical natural language processing (cNLP), rule-based approaches still have a prominent role. 
In this paper, we introduce medspaCy, an extensible, open-source cNLP library based on spaCy framework that allows flexible integration of rule-based and machine learning-based algorithms adapted to clinical text. MedspaCy includes a variety of components that meet common cNLP needs such as context analysis and mapping to standard terminologies. By utilizing spaCy's clear and easy-to-use conventions, medspaCy enables development of custom pipelines that integrate easily with other spaCy-based modules. Our toolkit includes several core components and facilitates rapid development of pipelines for clinical text. 
}

\section*{Introduction}
Retrospective clinical research often relies on data extracted from electronic medical record (EMR) systems using natural language processing (NLP) adapted for clinical language.  Despite a wide range of existing solutions, code reuse is challenging  because new system development projects face the labor-intensive task of connecting isolated modules into cohesive cNLP pipelines. 
Over the years, the need for reuse and reproducibility has been met with a number of frameworks and toolkits.\cite{Digan2021CanSuites} Java-based general architectures, such as UIMA \cite{FERRUCCI2004} and GATE \cite{Cunningham2002GATEEngineering}, have provided a strong foundation for a number of highly successful cNLP toolkits and general purpose reusable systems such as cTAKES, CLAMP,  Leo, and HITEx \cite{Savova2010,Soysal2018,Cornia2014a,Zeng2006ExtractingSystem}. As a platform-independent and straightforward programming language, Java has been the primary environment for most cNLP applications of the last 20 years even prompting re-implementation of tools previously written in other programming languages to improve accessibility \cite{Demner-Fushman2017,Aronson2001}. 

While Python programming language is celebrating its 30-year anniversary, in the last few years its usage has exploded as it had become one of the most popular languages of data science.\cite{Vizard2021PythonLanguage, Hayes2020Usage,Piatetsky2019,Github2020TheOctoverse,20202020Survey} The wider NLP, machine learning, and data science community's shift towards Python programming language is fueled by its capability of interactive environments to explore data and experiment with approaches while avoiding a common cycle of compiling code and reloading data \cite{Perez2007}. 
The machine learning ecosystem in Python is vibrant and growing, including state-of-the-art training methods and deep learning models.

Until recently Natural Language Toolkit (NLTK) has been the leading platform for general text processing in Python. \cite{Bird2009} Actively supported by a dedicated team since 2000, NLTK is comprised of a number highly functional NLP libraries for rule-based and machine learning-based text analysis. While NLTK has enabled hundreds of successful research and educational projects, its approach to text processing as lists of strings and lack of a unifying architecture have hindered its ability to scale to large datasets.  

In contrast, spaCy\footnote{https://spacy.io/} provides a robust architecture for building and sharing custom, high-performance NLP pipelines by taking an object-oriented view of text. It is non-destructive, supports seamless integration of statistical and machine-learning methods with rule-based NLP, and allows for the creation of custom components for specialized tasks. Powered by the strength of Cython, an optimizing static compiler for Python that generates very efficient C or C++ code, spaCy allows achieving exceptional speed of performance. Similar to UIMA approach in Java, spaCy provides a framework for modular plug-and-play construction of custom NLP pipelines. Since its inception in 2015, spaCy universe has gained a robust, highly-active, and growing community of contributors of open source modules, state-of-the-art models, and end-to-end systems developed with or for the framework.

Recognizing the need to adapt processing to different domains, several models and toolkits have been introduced targeting specialized text, among them scispaCy that handles scientific and biomedical sublanguages \cite{Neumann2019}. While scispaCy includes custom tokenization rules, biomedical concept identification, and a variety of pre-trained dependency parsing and named entity recognition (NER) models, the differences between clinical and biomedical language hamper its ability to achieve optimal performance on  clinical narrative. To confront the particular challenges posed by medical text, a team at Virginia Commonwealth University created MedaCy, a medical text mining framework built over spaCy to facilitate the engineering, training and application of machine learning models for medical information extraction \cite{Mulyar2018}. Neither scispaCy nor MedaCy provide adequate support for rule-based system development beyond the functionality natively provided by spaCy. Despite a wide range of machine learning applications, rule-based algorithms still comprise a large proportion of successful academic and commercial cNLP projects
\cite{Fu2020ClinicalReview,Sheikhalishahi2019NaturalReview.}. 

To bridge the gap between the existing functionality and the needs of the clinical research community for easy-to-use clinical text processing that combines statistical and symbolic methods, a team of enthusiastic cNLP practitioners at the Veterans Health Administration and University of Utah developed a spaCy-based library of core components targeting medical text called medspaCy.

\section*{Methods}

\subsection*{Framework Overview}

\begin{figure}[t]
\centering
\vspace{-5pt}
\includegraphics[scale=.5]{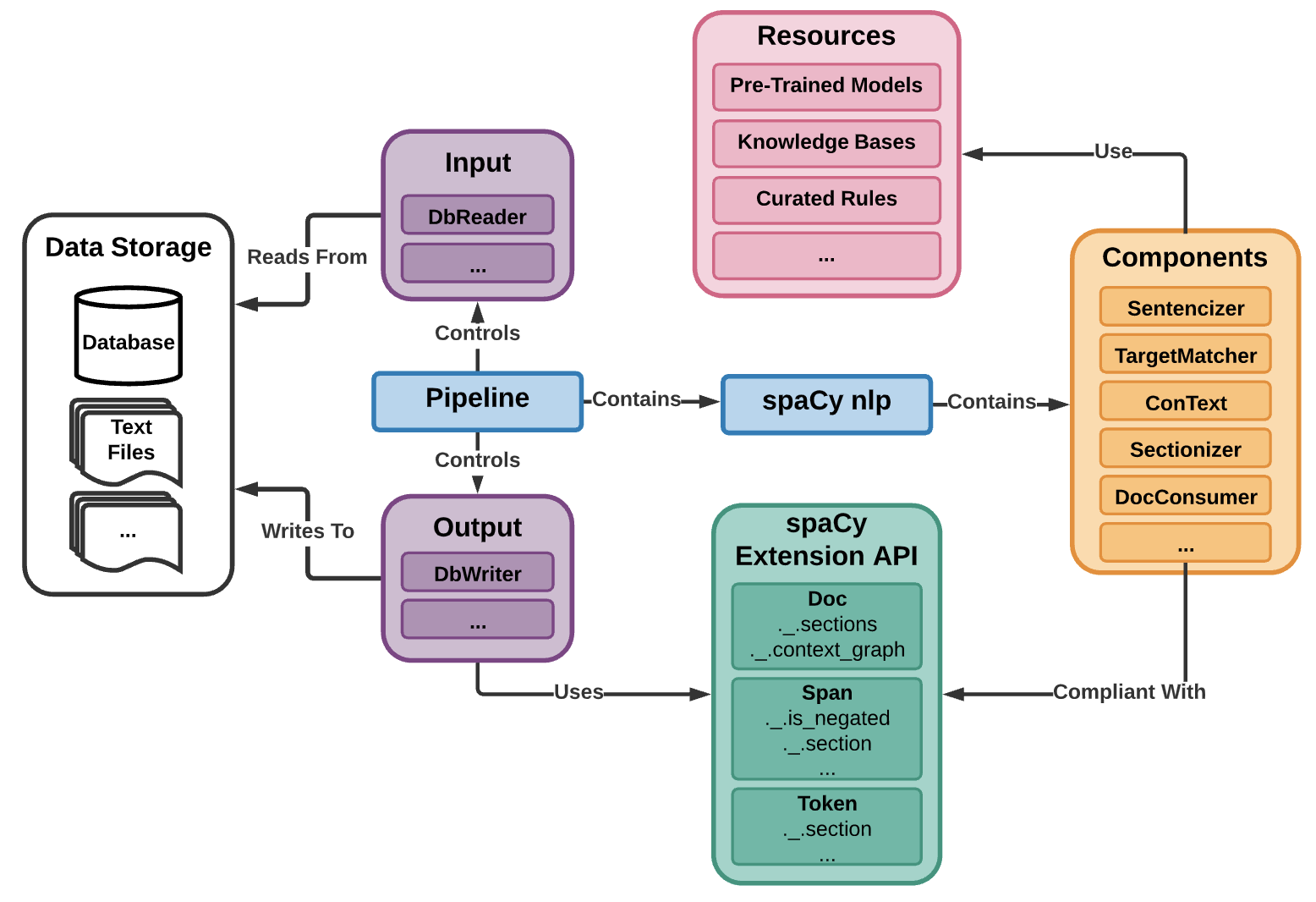}
\vspace{-5pt}
\caption{Overview of medspaCy architecture.}
\label{fig:medspacy1}
\end{figure}

MedspaCy is designed to utilize spaCy's application programming interface (API) while only minimally expanding upon it where needed. This minimal expansion of spaCy's API ensures each component is usable standalone or as one part of a larger spaCy pipeline that may incorporate components from other sources. Figure \ref{fig:medspacy1} illustrates the interconnection of medspaCy components into an end-to-end system through a common API. Once the toolkit reached acceptable maturity, we made it available through PyPI with the command \texttt{pip install medspacy}.

All information generated by components is directly accessible through the use of spaCy's attribute extension functionality. For example, a default spaCy attribute of a named entity is accessible using \texttt{ent.like\_num} attribute, a medspaCy attribute utilizes the custom extensions such as \texttt{ent.\_.is\_negated}. Attribute extensions are registered at the \texttt{Doc}, \texttt{Span}, or \texttt{Token} levels, depending on the component to create a variety of useful tools for analyzing data after medspaCy processing is complete.

\begin{figure}[t]
\centering 
\includegraphics[scale=.25]{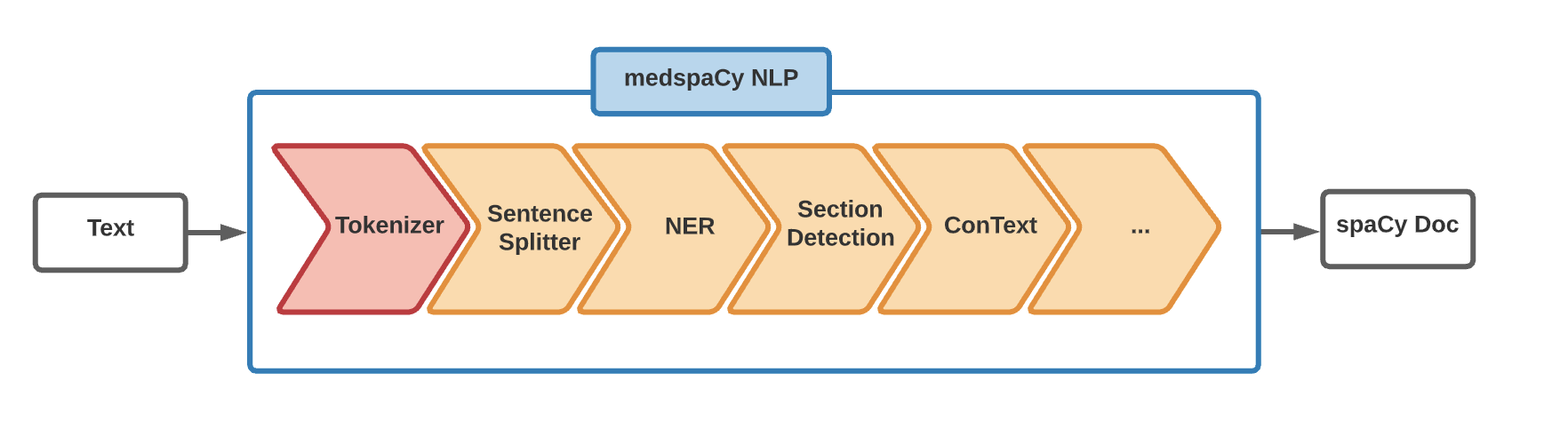}
\vspace{-5pt}
\caption{Example of a text processing pipeline that utilizes medspaCy and other spaCy-based components.}
\label{fig:medspacy2}
\end{figure}

The components included in medspaCy offer both a default initialization and a large number of optional customization parameters. This allows components to be quick to set up, learn, and use to develop prototypes while also being fully customizable for more specific needs. Resources, such as curated rule sets or knowledge bases, may be shared with the community and utilized across projects.
Figure \ref{fig:medspacy2} is an illustration of an example pipeline using medspaCy. Since the spaCy API allows for modular components, there are many possible pipelines where users could insert alternative components, such as scispaCy AbbreviationDetector or other spaCy-compatible rule-based or machine learning-based modules.

\subsection*{MedspaCy Components}
MedspaCy has a growing list of integrated components that allow building end-to-end systems for clinical text processing.

\subsubsection*{Tokenization}
The primary advantage of spaCy's approach to tokenization is that it is non-destructive, which preserves all whitespace and punctuation information enabling complete reconstruction of the original text. The major drawback of the default spaCy tokenizer for clinical text processing is that it is not trained on clinical text. Additionally, it has a variety of rules designed to handle text sourced online, including many rules that mitigate excess tokenization of URLs. These rules prevent splitting sequences of alphanumeric characters and punctuation into multiple tokens. 
However, URLs are relatively uncommon in clinical text but typos and using punctuation to delineate document structures are common. The tokenizer included in medspaCy implemented custom rules to handle punctuation and inconsistent use of whitespace that are common in clinical notes. 

\subsubsection*{Sentence Detection}
Due to the limited purpose of clinical language to document and succinctly communicate information about a specific patient, sentences in clinical text are characterized by telegraphic grammar with omitted subjects and the propensity to use long lists for present or absent conditions and prescribed medications \cite{Ford2016}. Sentence segmentation in clinical language is hindered by frequent use of templated and tabular text, abundance of concept-value pair statements, and inconsistent use of punctuation.  Since the text is meant to be read within a specific user interface of the EMR, space, tab, and new line characters are utilized to indicate sentence or section boundaries as well as for visual indication of grouped text, such as in case of templates and tabular text. Syntactically significant whitespace complicates the traditional approach to text processing that ignores existence and length of whitespace in text.
To correctly identify the context of target concepts from clinical text, a clinical-domain sentence boundary detector is essential, because many downstream components work at the sentence level. Commonly used sentence detectors trained on non-clinical text typically do not perform well on clinical text data.\cite{Shi2016} To make medspaCy more capable of performing cNLP, we adopted a high-performance, rule-based sentence detector for clinical text -- PyRuSH, which is the Python version of Rule-based sentence Segmenter using Hashing (RuSH)\cite{Shi2016}. Relying on rules for tokenization allows easy adaptation of the tokenizer to a new setting without extensive retraining.

\subsubsection*{Section Detection}
Clinical documentation has a well established data format for different document types. Logical Observation Identifiers Names and Codes (LOINC) Document Ontology outlines hundreds of document types that vary by subject domains, role of the author, setting, type of service described, and document kind \cite{Hyun2006}. At each intersection of these axes, a clinical document has a specific structure that is either learned through apprenticeship (medical students learning from practice) or specifically prescribed by guidelines \cite{Goldsmith2008, Zeng2011}. The same statement in different sections of the same or different documents might have a very different meanings. For example, a document type called \emph{Progress Note} has Subjective, Objective, Assessment, Plan sections, and \emph{Procedure Report} has Findings and Impressions sections. Depending on which section it is mentioned in, a clinical condition may be experienced by a patient in the past, may be happening at the time when the note is written, or the patient may be at risk for experiencing it in the future. These sections may or may not be explicitly labeled with a variety of section headers fully spelled out or abbreviated. 

Breaking documents into relevant sections is often a core part of cNLP, especially as documents grow in length and complexity. medspaCy includes an implementation of clinical section detection based on rule-based matching of the section titles with default rules adapted from SecTag\cite{Denny2008} and expanded through practice. The sectionizer contains several other features beyond simple section title identification. If desired, the sectionizer can interact with the attributes of entities, such as changing the temporality attribute of all entities in the ``Past Medical History" section. The medspaCy sectionizer is also capable of creating hierarchies of sections and subsections within the document and preserving them in a traversable tree structure.

\subsubsection*{Concept Extraction}
Clinical concept extraction through concept mention detection and concept encoding is one of the most prominent tasks of clinical text processing. \cite{Fu2020ClinicalReview}
Similar to NER, concept extraction employs either manual rules or trained statistical models to identify specific spans of text from a clinical document and labeling them with pre-defined clinical concepts. Several utilities are included in medspaCy for targeted concept extraction which extend existing spaCy rule-based functionality. Rules defining concepts specify the text pattern to be matched, semantic category, and additional metadata about a concept. In addition to integrating existing pattern-matching functionality provided by spaCy, medspaCy incorporates additional regular expression features, such as accepting multi-token regular expressions, and allows directly specifying entity attributes, such as temporality or standard vocabulary codes. Enabling additional regular expression matching ensures greater compatibility with existing cNLP resources and knowledge bases as regular expressions are commonly used in other frameworks and languages.

\subsubsection*{UMLS Mapping}
Clinical text processing systems frequently rely on comprehensive indexing of the narrative text to a standard vocabulary. 
The Unified Medical Language System (UMLS) \cite{Bodenreider2004} is the most comprehensive standard terminology that is typically used as the basis for clinical concept extraction. Aiming to enable UMLS concept extraction with minimal environment configuration, medspaCy adapted QuickUMLS to spaCy framework. The existing system QuickUMLS was selected as a concept mapping solution due to its speed and matching accuracy that are comparable to other existing systems \cite{Soldaini2016 }. It allows efficient approximate dictionary matching by leveraging an implementation of the SimString library. \cite{Okazaki2010} Modifications were made to QuickUMLS so that its algorithm could be contained in a spaCy component.

UMLS licensing does not allow redistribution of the databases, therefore, medspaCy is distributed with resources generated from a publicly available UMLS sample.\footnote{\href{https://www.nlm.nih.gov/research/umls/new_users/online_learning/Meta_006.html}{https://www.nlm.nih.gov/research/umls/new\_users/online\_learning/Meta\_006.html}} MedspaCy documentation includes instructions on how to create QuickUMLS resources after installation. An additional contribution of medspaCy is a better support for QuickUMLS and its dependencies for Windows operating system. While our implementation of QuickUMLS in Windows requires Anaconda and some additional installation steps, other operating systems such as Linux and MacOS can perform extractions with the provided small UMLS sample immediately following installation. 

\subsubsection*{Contextual Analysis}
Extracting contextual properties of clinical concepts is an essential step of typical cNLP systems. Negation, temporality, certainty, and experiencer are the most useful contextual features for entities extracted from clinical text. The ConText algorithm\cite{Harkema2009a} asserts attributes by linking entities with linguistic modifiers within a contextual window, typically a sentence. 
Previously implemented using Python in pyConTextNLP \cite{Chapman2011}, the algorithm was adapted to spaCy API with extended attributes. 
MedspaCy implements the ConText algorithm with a set of default rules that can be customized as needed. ConText rules take advantage of pattern matching and also allow the user to control the behavior of the modifier by defining properties such as directionality in the text and custom linking logic. Several attributes, such as negation, are registered by default, but attributes can be customized by users to allow for use case-specific semantic categories.

\subsubsection*{Utilities}
Assisting with development of complete end-to-end systems, medspaCy provides a variety of utilities to support deploying pipelines and analyzing output of text processing: \begin{itemize}
\item pre- and post-processing utilities that help developers transform simple methods into components compatible with spaCy pipelines.
\item a component that converts a spaCy \texttt{Doc} to a dictionary-based format that is directly convertible to Pandas, SQL, CSV, or other tabular format when added to the end of an existing medspaCy pipeline. It supports output for all attributes and spans for entities, sections, and context.
\item a pipeline controller for deploying medspaCy on large quantities of data. It facilitates batching documents on both input and output and a wrapper that handles standard database operations through \texttt{pyodbc} (Python ODBC bridge), such as creating tables and write queries given a set of desired columns, which allows medspaCy to deploy faster.
\end{itemize}

\subsubsection*{Visualization}
\begin{figure}[h]
\centering 
\includegraphics[width=0.55\textwidth]{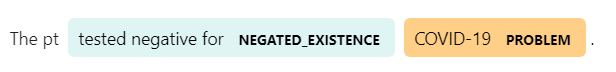}
\\
\includegraphics[width=0.55\textwidth]{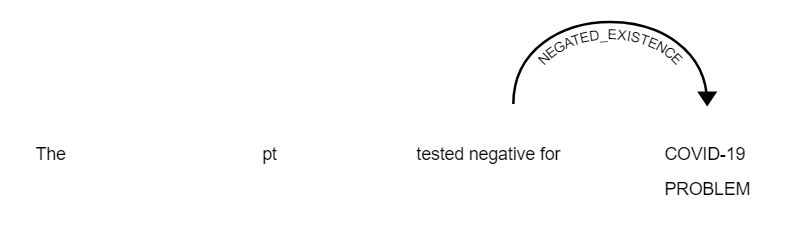}
\caption{Example of visualization using highlight and context arrows.}
\label{fig:medspacy3}
\end{figure}

The visualization tools included in spaCy's have been adapted to display the results of the integrated medspaCy components. There are two types of visualization included (Figure \ref{fig:medspacy3}): the first highlights extracted entities, contextual modifiers, and section titles in the text. The second represents the ConText algorithm by drawing directed arrows between entities and their linked modifiers.
These visualizations can be used to present a model's final output or to explain the logic of a model. During development, visualizing intermediate results can assist with debugging. Updating a model, testing on text, and inspecting the results with the visualizer can be done rapidly when used in an interactive setting such as Jupyter Notebook.

\section*{Use Cases}

Multiple operational and research oriented projects within the U.S. Department of Veterans Affairs (VA) have utilized medspaCy. 
The rapid development supported by medspaCy has been essential to several projects aiding the VA's COVID-19 pandemic response.

\subsection*{Chief Complaint Surveillance} 
One of the earliest applications utilizing some of these components in a spaCy pipeline is a system which is used for enhanced syndromic surveillance in VA. This system processes presenting symptoms recorded during emergency department triage such as \texttt{"n/v/d"} (i.e. ``nausea / vomiting / diarrhea") or \texttt{"c/o cp+sob"} (i.e. ``complains of chest pain and shortness of breath") to extract UMLS concepts. This pipeline has been running in an operational capacity for biosurveillance since early 2019. It was leveraged in early 2020 in response to the COVID-19 pandemic to identify potential patients under investigation before the existence of laboratory testing for SARS-CoV-2 or the term COVID-19 was adopted. This pipeline performs text pre-processing using a subset of regular expression patterns from the Emergency Medical Text Preprocessor (EMT-P) \cite{Travers2004}, and maps text spans to standard vocabulary concepts using the QuickUMLS component.

\subsection*{COVID-19 Surveillance} 
Another operational project used medspaCy to identify patients who have were tested for COVID-19 outside of the VA network when no structured lab results were available \cite{Chapman2020}. This pipeline classifies clinical documents which likely contain a mention of positive COVID testing so that they can be prioritized for manual chart review. Between January 1 and June 15, 2020 the system had processed documents for over 3 million patients and had resulted in over 36\% of all identified COVID positive patients being identified by this method alone. An evaluation of the system demonstrated precision and recall of 82.5\% and 94.2\%, respectively. As terminology related to the virus changed over time, it was imperative to add to concepts and patterns nearly every day. Iterative development of this system with medspaCy permitted putting the system into practice by January 21, 2020 and has been continuously operational throughout the pandemic, having processed over 63 million clinical documents one year after initial deployment. 
\footnote{https://spacy.io/universe/project/cov-bsv}

\subsection*{VA COVID-19 Screening}
As the COVID-19 pandemic spread, VA facilities implemented standardized COVID-19 screenings for all visitors and employees. All interactions with Veterans are logged using a specific document format that quickly became one of the most frequent documents entered into the VA EMR in 2020. Veterans in inpatient care or community living centers often had multiple COVID-19 screenings recorded each day.

A medspaCy pipeline was developed to identify the screening as a standalone document or as a subsection of a larger document and then to extract up to 17 different possible screening questions and the associated responses including community- or travel-based exposure to COVID-19, previous COVID-19 tests or diagnoses, or specific COVID-19 symptoms.

This pipeline has processed 6.8 million screening documents for 914,000 Veterans who have received a COVID-19 test. The dataset is available to VA researchers in the VA COVID-19 Shared Data Resource. 

\subsection*{Inpatient Nursing Templated Text Processing}
Several national VA standard operating procedures have been put into operation during the COVID-19 pandemic mandating use of standardized inpatient nursing shift assessments. While the assessments are presented to users as forms, the data are stored as the raw text conversions of templated questionnaires with their answers, thus, much of the information is not automatically retrievable. The medspaCy sectionizer is being utilized for the automatic extraction of information from these semi-structured documents. Given an empty version of each template, a utility module is used to automatically generate sectionizer rules. Sectionizer applies those rules to identify each question in the template as section header and extracts of all template answers captured as the section body.

\begin{figure}[h]
\centering 
\includegraphics[scale=.5]{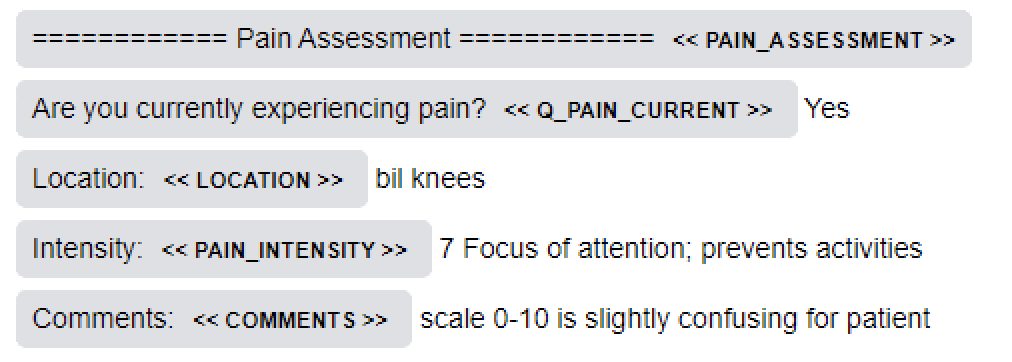}
\caption{Highlighted section headers delineate template questions with the section body as the responses when processing semi-structured text.}
\label{fig:medspacy4}
\end{figure}


\subsection*{Veteran Suicide Risk and Intervention Analyses} 
Several capabilities of medspaCy were used to rapidly summarize insights from Veterans, family members and organizations who responded to a request for information on how to improve suicide prevention. This survey was performed as part of the President’s Roadmap to Empower Veterans and End a National Tragedy of Suicide (PREVENTS) task force to better understand risk factors and approaches to end Veteran suicide \cite{USDepartmentofVeteransAffairs2020}. A total of 9,040 open ended responses needed to be analyzed and their insights summarized to allow a report of findings to be delivered to the task force within 3 months. Several capabilities in medspaCy made this project possible within the requested deadline, such as iterative rule development, exploratory data analysis, and visualization of text extraction. The project was also able to leverage existing knowledge resources from UMLS, including concepts related to mental health, medication and treatment. These insights were utilized by the task force as material to inform development of the PREVENTS\cite{USDepartmentofVeteransAffairs2020}.

\subsection*{Homelessness and Housing Stability Identification} 
Social determinants of health including homelessness are important factors in patient health and are often only documented in free text \cite{Conway2019Moonstone:Narratives,Gundlapalli2014ExtractingRecords}. The VA partners with community organizations through the Supportive Services for Veteran Families (SSVF) program to provide rapid rehousing and temporary financial support to veterans who are homeless or at risk of becoming homeless \cite{Nelson2021AssociationProgram}. A pipeline is being developed using medspaCy which extracts mentions of homelessness and housing stability from clinical texts. This information is then aggregated to infer a patient's housing stability over a period of time and will be used to evaluate long-term outcomes of SSVF participants and identify at-risk individuals who could benefit from enrollment in the program.

\subsection*{Education} 
Besides supporting operational or clinical application, medspaCy's simple interface and low barrier to entry makes it useful for education. Local, national, and international workshops and tutorials have utilized medspaCy to illustrate clinical text processing. A recent multi-day intensive course on clinical data science targeted to medical students used medspaCy as part of its curriculum\footnote{\href{https://github.com/Melbourne-BMDS/mimic34md2020_materials}{https://github.com/Melbourne-BMDS/mimic34md2020\_materials}}. This course introduces medspaCy one one of the first sessions so that the remainder of the workshop can be devoted to experiential learning where students create rules and modify components to develop systems in processing clinical documentation. A simple installation process and a common programming interface enables students with little programming experience to quickly apply NLP to clinical data.


\section*{Future work}
There are many additional components and applications possible in the next steps for medspaCy. One potential area of development is releasing medspaCy pipelines which include components, resources, and pre-trained models  for specific cNLP tasks. These could also be created from publicly available resources such as MIMIC-III \cite{Johnson2016}, shared tasks such as i2b2 \cite{Sun2013EvaluatingChallenge}, or ontologies constructed during previous research. Released models could include a broad range of clinical domains, including adverse drug event detection \cite{Chapman2019,Henry}, infectious disease surveillance, and radiology reporting.

While most of the existing medspaCy components support rule-based systems, future work could include additional utilities for machine learning. This could include word embeddings loaded as part of a default pipeline or utilities for feature extraction from text. Such improvements would allow users to more fully leverage both statistical and rule-based cNLP.

\section*{Conclusion}
By improving the accessibility to high-performance NLP solutions optimized for clinical language, we aim to improve the productivity of cNLP development and reduce development burden for projects that require the use of rule-based NLP or the integration of rules with other NLP methods. Unlocking new data elements through clinical information extraction will enable researchers to find answers to questions previously inaccessible for investigation.

\section*{Acknowledgements}
This work was supported using resources and facilities of the Department of Veterans Affairs (VA) Informatics and Computing Infrastructure (VINCI), VA HSR RES 13-457, the Veterans Health Administration (VHA) Office of Analytics and Performance Integration (API), BASIC (Biosurveillance, Antimicrobial Stewardship, and Infection Control) and 
by the VA HSR\&D Informatics, Decision-Enhancement, and Analytic Sciences (IDEAS) Center of Innovation (CIN 13-414).

The views expressed in this article are those of the authors and do not necessarily reflect the position or policy of the Department of Veterans Affairs or the United States government.
 
\section*{\centering References}
\bibliographystyle{vancouver}
\renewcommand{\section}[2]{}%
\bibliography{references}
\end{document}